\documentclass{article}

\usepackage[preprint, nonatbib]{neurips_2023}

\usepackage[utf8]{inputenc} 
\usepackage[T1]{fontenc}    
\usepackage[colorlinks,
            linkcolor=BrickRed,       
            anchorcolor=BrickRed,  
            citecolor=BrickRed,       
            ]{hyperref}       
\usepackage{hyperref}       
\usepackage{url}            
\usepackage{booktabs}       
\usepackage{nicefrac}       
\usepackage{microtype}      

\usepackage{graphicx}
\usepackage[caption=false,font=normalsize,labelfont=sf,textfont=sf]{subfig}
\usepackage[table, svgnames, dvipsnames]{xcolor}
\usepackage{amsmath,amssymb,amsfonts}
\usepackage{multirow}
\usepackage{makecell}
\usepackage{ulem}
\usepackage{pifont}
\usepackage{multirow}
\usepackage{makecell}
\usepackage{wrapfig}
\usepackage{algorithm}
\usepackage{algorithmic}
\usepackage{soul}
\usepackage{float}
\usepackage{arydshln}
\usepackage{xcolor}
\newcommand{\bmmc}[1]{\bm{\mathcal{#1}}}

\newcommand{\bm}[1]{\boldsymbol{\mathcal{#1}}}
\DeclareMathOperator*{\argmax}{arg\,max}

\title{Reconstruction Distortion of Learned Image Compression with Imperceptible Perturbations}

\author{%
  Yang Sui \thanks{The work of Y. Sui was done during the internship at Tencent America.} \\
  Rutgers University \\
  \And
  Zhuohang Li \thanks{The work of Z. Li was done during the visit at Tencent America.} \\
  Vanderbilt University \\
    \And
    Ding Ding\\
    Tencent America\\
    \And
    Xiang Pan\\
    Tencent America\\
    \And
    Xiaozhong Xu\\
    Tencent America\\
    \And
    Shan Liu\\
    Tencent America\\
    \And
    Zhenzhong Chen \thanks{The work of Z. Chen was done during the visit at Tencent.}\\
    Wuhan University\\
}

\begin{document}


\maketitle

\begin{abstract}
    Learned Image Compression (LIC) has recently become the trending technique for image transmission due to its notable performance. Despite its popularity, the robustness of LIC with respect to the quality of image reconstruction remains under-explored. In this paper, we introduce an imperceptible attack approach designed to effectively degrade the reconstruction quality of LIC, resulting in the reconstructed image being severely disrupted by noise where any object in the reconstructed images is virtually impossible. More specifically, we generate adversarial examples by introducing a Frobenius norm-based loss function to maximize the discrepancy between original images and reconstructed adversarial examples. Further, leveraging the insensitivity of high-frequency components to human vision, we introduce Imperceptibility Constraint (IC) to ensure that the perturbations remain inconspicuous. Experiments conducted on the Kodak dataset using various LIC models demonstrate effectiveness. In addition, we provide several findings and suggestions for designing future defenses.
\end{abstract}


\section{Introduction}

Learned Image Compression (LIC) has recently gained tremendous success in transmitting images under bit-rate limits because of its superior performance over traditional image compression. Specifically, LIC frameworks \cite{balle2017endtoend, balle2018variational, minnen2018joint, cheng2020learned} exhibit significant rate-distortion (R-D) performance and outperform standard image compression methods such as JPEG \cite{wallace1991jpeg}, JPEG2000 \cite{taubman2002jpeg2000}, and BPG \cite{bellard2016bpg}.
Generally, the fundamental LIC structure leverages the auto-encoder framework, which adopts a Deep Neural Network (DNN)-based non-linear transformation with an encoder for image compression and a decoder for reconstruction. 

Despite demonstrating a high recovery ability, the DNN-based encoder and decoder also bring concern about robustness. Adversarial attacks~\cite{szegedy2014intriguing,goodfellow2015explaining} are a type of security threat against DNN-based systems where the attacker injects specially crafted perturbations to images to construct adversarial examples, which are natural-looking images that can cause misclassification to DNN models. Originally discovered in image classification tasks~\cite{szegedy2014intriguing, goodfellow2015explaining}, adversarial examples have then attracted great research attention in many fields of computer vision, including object detection and semantic segmentation~\cite{xie2017adversarial}, facial recognition~\cite{dong2019efficient}, and visual question answering~\cite{li2021adversarial}. 

While the robustness of downstream tasks has been extensively investigated~\cite{carlini2017towards,madry2018towards}, the robustness of LIC, which is evaluated in terms of image reconstruction quality instead of classification accuracy, has received little attention. As depicted in Fig. \ref{fig:pipeline}, a typical LIC system can accurately reconstruct an original (unperturbed) image.
However, the LIC is considered to be not robust if an attacker can introduce small perturbations into the original image to significantly disrupt the reconstructed image, resulting in the main object in the reconstructed image being unrecognizable.

In this paper, we propose to investigate the robustness of LIC by attacking the image reconstruction process. The main idea is to solve an optimization problem to minimize the adversarial perturbation while maximizing the \textbf{\textit{Frobenius norm-based loss}} metric between the original and reconstructed images. However, the adversarial perturbations generated with unconstrained Frobenius norm-based loss are always sensitive to human eyes. To improve the imperceptibility of the generated adversarial images, we draw insights from the observation that high-frequency components are less perceptible to human vision \cite{sharma2019effectiveness}, and consider generating perturbations from a frequency perspective by introducing a Discrete Cosine Transform (DCT)-based \textbf{\textit{Imperceptibility Constraint (IC)}} into the adversarial loss function, rendering the perturbations more unnoticeable by human perception. Our contributions can be summarized as follows: \ding{182} We conduct a systematic investigation on the robustness of LIC by launching a series of attacks that disrupts the image reconstruction process by introducing a Frobenius norm-based loss with IC. 
\ding{183} Our experiments demonstrate that our proposed attack can disrupt LIC while maintaining the imperceptibility of the induced perturbations.
\ding{184} Based on our experiments, we provide several intriguing findings and potential insights regarding designing robust LICs.

\section{Preliminary}

\subsection{Learned Image Compression}
\label{method:LIC}


Given represent the non-linear encoder $g_a(\cdot)$ and decoder $g_s(\cdot)$, let $\bmmc{X}$ and $\bmmc{\hat{X}}$ denote the original input and reconstructed images, and $\bmmc{Y}$ and $\bmmc{\hat{Y}}$ denote the pre-quantized and quantized latent representation, respectively. The image compression process is formulated as follows:
\begin{equation}
\begin{aligned}
\bmmc{Y} = g_a (\bmmc{X}), \enspace \bmmc{\hat{Y}} = \text{AD}(\text{AE}(Q(\bmmc{Y}))), \enspace \bmmc{\hat{X}} = g_s(\bmmc{\hat{Y}}),\\
\end{aligned}
\label{eqn:lic}
\end{equation}
where $Q(\cdot)$ is the quantization operation, and $\text{AE}$ and $\text{AD}$ represent the arithmetic encoding and decoding processes, respectively. The reconstructed image $\bmmc{\hat{X}}$ is the output of the corresponding (inverse) transform. In addition, a hyper-prior is used as side information to reduce the bit-rate. 

\subsection{Adversarial Attack}

Given a natural image $\bmmc{X} \in \mathbb{R}^{H \times W \times C}$, the corresponding label $k$, and a classification model $f(\cdot)_{i}$ which predicts the probability of the image belonging to class $i$, the goal of adversarial attacks is to craft an adversarial perturbation $\bmmc{\delta} \in \mathbb{R}^{H \times W \times C}$ to be added onto $\bmmc{X}$ so that it is misclassified by $f(\cdot)$, which can be formulated as:
\begin{equation}
\begin{aligned}
\argmax \limits_{i} f(\bmmc{X}+\bmmc{\delta})_{i} \neq k, \quad \| \bmmc{\delta} \|_{p} \leq \epsilon, \\
\end{aligned}
\label{eqn:adv_attack}
\end{equation}
where $\epsilon$ controls the perturbation budget. 

\section{Distortion with Imperceptible Perturbation}

\begin{figure}[t]
\centering
  \includegraphics[width=0.5\linewidth]{ 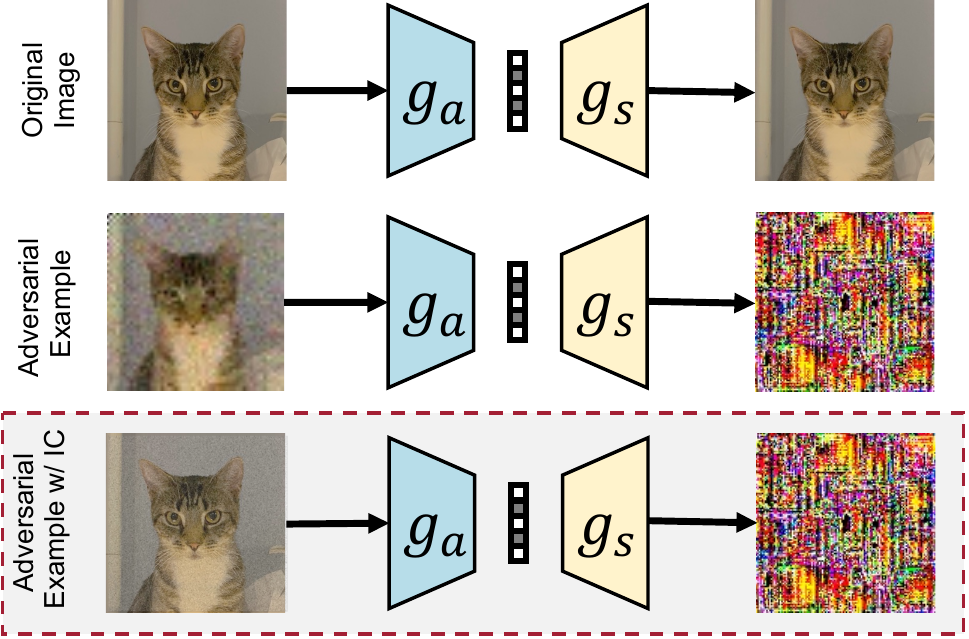}
\caption{Illustration of proposed adversarial attack with IC on LIC to disrupt the reconstruction image. Top: The standard LIC process. Middle: The proposed adversarial attack against LIC. Bottom: The proposed adversarial attack with IC against LIC while ensuring the perturbation on the original image remains more imperceptible.}
\label{fig:pipeline}
\end{figure}

\subsection{Reconstruction Distortion Attack}

Unlike the traditional adversarial attack, which aims to mislead the classification model into predicting a wrong label, the adversarial attack on LIC aims to introduce small noise to the original image so that the reconstructed image is severely corrupted.
This objective can be formulated as:
\begin{equation}
\begin{aligned}
\argmax \limits_{\bmmc{\delta}} \enspace \texttt{dis}(\bmmc{X}, g_s(Q(g_a((\bmmc{X}+\bmmc{\delta}))))), \enspace \| \bmmc{\delta} \|_{p} \leq \epsilon, \\
\end{aligned}
\label{eqn:distortion_attack}
\end{equation}
where $\texttt{dis}(\cdot, \cdot)$ denotes a distance function that computes the distance between two tensors. In addition, we utilize differential approximation quantization \cite{shin2017jpeg} to achieve gradient-based attacks. We note that, in practice, the underlying technology for image compression is often standardized or industrially recommended (such as JPEG compression) to ensure compatibility across all scenarios.
Therefore, we assume the attacker has complete knowledge of the LIC and can launch the attack in a white-box manner.

\subsection{Imperceptible Perturbations}
As illustrated in Fig. \ref{fig:pipeline} (middle), although the image can be heavily distorted by solving the Eq. \ref{eqn:distortion_attack},
in order to reduce the reconstructed image quality, a significant amount of perturbation needs to be injected, which makes the primary subject of the image barely recognizable and the attack easily detectable by human inspectors.

Previous research \cite{sharma2019effectiveness} has demonstrated that high-frequency perturbations are less noticeable. Typically, performing DCT on an image reveals that the low-frequency components carry the major semantic information, while the high-frequency components include the edge structural information. Human beings tend to focus on semantic information, such as the main object in an image, but ignore detailed information, particularly beneath edge structures. Hence, human visual perception is typically not sensitive to these high-frequency perturbations.

Motivated by the frequency perspective, we propose a DCT-based IC to generate imperceptible high-frequency perturbations. In particular, IC encourages the perturbation to mainly modify the high-frequency components of the original images, while constraining the low-frequency components of the adversarial images to remain consistent and close to those in the original images, which can be formulated as:
\begin{equation}
\begin{aligned}
\mathcal{I}(\bmmc{X}, \bmmc{X} + \bmmc{\delta}) = \| \mathcal{T}(\bmmc{X}) - \mathcal{T}(\bmmc{X} + \bmmc{\delta}) \|_F, 
\end{aligned}
\label{eqn:IC}
\end{equation}
where $\| \cdot \|_F$ denotes the Frobenius norm of the distance. $\mathcal{T}(\cdot)$ denotes the function truncates the low-frequency band based on DCT, which can be formulated as follows:
\begin{equation}
\begin{aligned}
\mathcal{T}(\bmmc{X}) = \texttt{IDCT}(\boldsymbol{M} \odot \texttt{DCT}(\bmmc{X})), 
\end{aligned}
\label{eqn:dct}
\end{equation}
where $\odot$ denotes the Hadamard product (element-wise product).
$\boldsymbol{M} \in \mathbb{R}^{H \times W}$ is a binary mask applied to the frequency domain of the tensor $\bmmc{X}$ after DCT to constrain its frequency components. To make the perturbation imperceptible, we mask out half of the components with lower frequencies and only preserve the higher half of the frequency components.

By combining Eq. \ref{eqn:distortion_attack} and Eq. \ref{eqn:IC}, the overall optimization objective is as follows:
\begin{equation}
\begin{aligned}
\mathcal{L}(\bmmc{X}, \boldsymbol{\delta}) =& -\| \bmmc{X} - g_s(Q(g_a((\bmmc{X}+\bmmc{\delta})))) \|_F^2 \\
& + \eta \cdot \mathcal{I}(\bmmc{X}, \bmmc{X} + \bmmc{\delta}), \quad \textrm{s.t.} \| \bmmc{\delta} \|_{\infty} \leq \epsilon,
\end{aligned}
\label{eqn:loss}
\end{equation}
where $\eta$ controls the influence of IC.

\section{Experiments}

\textbf{Setting.} For LIC models, we train the \texttt{Anchor} \cite{cheng2020learned} , \texttt{Hyperprior} \cite{balle2018variational}, \texttt{Factorized} \cite{balle2018variational}, and \texttt{Joint} \cite{minnen2018joint} LIC models following the CompressAI \cite{begaint2020compressai}, to evaluate the distortion. We use the differential approximation quantization from \cite{shin2017jpeg} to execute the gradient-based attack. We set $\eta=10^{2}$ for solving Eq. \ref{eqn:loss}.
The evaluations are conducted on the Kodak dataset using a single NVIDIA A100. 

\textbf{Metric.}
The effectiveness of the proposed attack is evaluated by Peak Signal-to-Noise Ratio (PSNR) and Multi-Scale Structural Similarity Index (MS-SSIM). A higher value of PSNR and MS-SSIM indicates better reconstruction quality. ``$\downarrow$ PSNR" and ``$\downarrow $ MS-SSIM" denote the decrease in PSNR and MS-SSIM, measured between adversarial examples and their reconstructed counterparts, where a higher value indicates a more effective attack.

\begin{figure*}[t]
	\centering 
	\includegraphics[width=\linewidth]{ 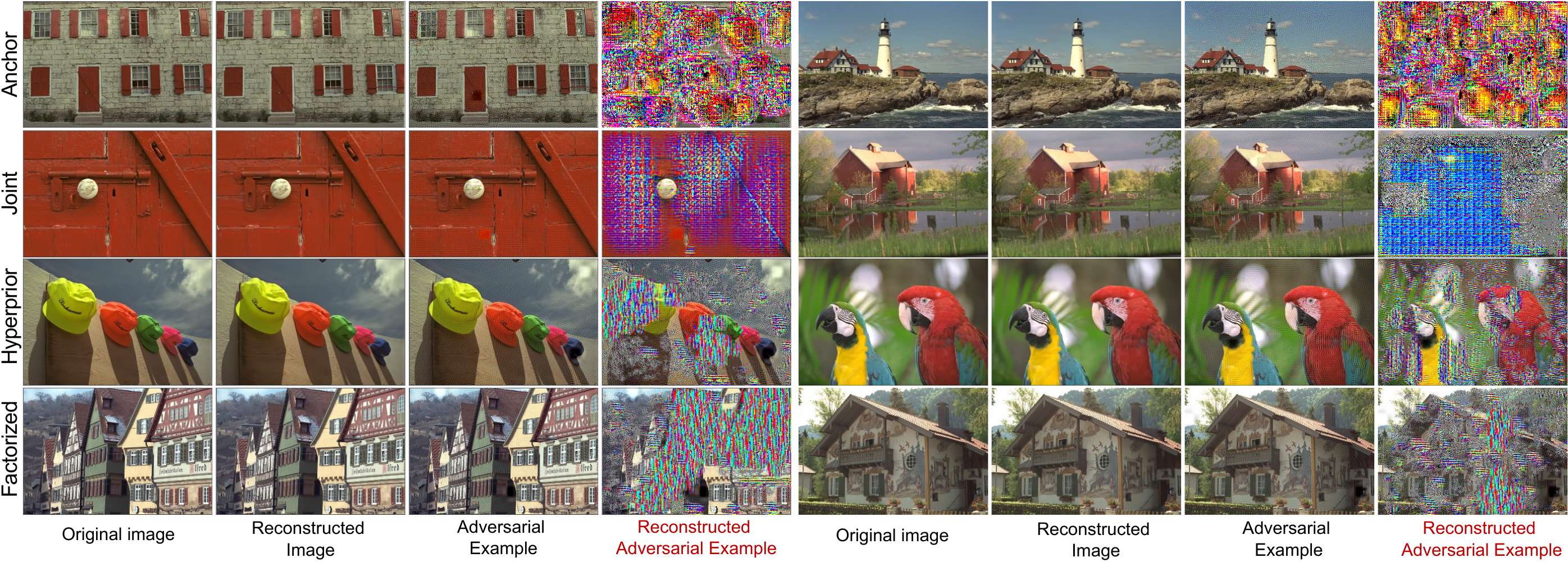}
	\caption{Visual results of our proposed reconstruction quality attack with IC. The attack is evaluated on the Kodak dataset with four various LIC models as victim models.}
	\label{fig:main_result}
\end{figure*}
\begin{table*}[ht]
\setlength{\tabcolsep}{1pt}
  \centering
  \scalebox{0.85}{
      \begin{tabular}{cccccccccc}
        \toprule
       \multirow{3}{*}{\makecell{ Model-$\epsilon$ }} & & \multicolumn{2}{c}{Reconstructed Original Image} & \multicolumn{2}{c}{AE} & \multicolumn{2}{c}{Reconstructed AE} & \multicolumn{2}{c}{$\Delta$ of AE vs. Reconstructed AE} \\
        
        \cmidrule{3-4} \cmidrule{5-6} \cmidrule{7-8} \cmidrule{9-10}
        & & PSNR & MS-SSIM & PSNR & MS-SSIM & PSNR & MS-SSIM & $\downarrow$ PSNR & $\downarrow$ MS-SSIM \\
        \midrule
        Anchor-32 & & 32.99 & 0.9880 & 22.47 & 0.9112 & \uline{7.500} & \uline{0.2188} & \cellcolor{Gainsboro!60}\textbf{14.97} & \cellcolor{Gainsboro!60}\textbf{0.6923} \\
        Anchor-64 & & 32.99 & 0.9880 & 17.63 & 0.8281 & \uline{6.617} & \uline{0.1816} & \cellcolor{Gainsboro!60}\textbf{11.01} & \cellcolor{Gainsboro!60}\textbf{0.6464} \\
        
        Hyperprior-32 & & 36.48 & 0.9943 & 20.30 & 0.8919 & \uline{12.05} & \uline{0.6617} & \cellcolor{Gainsboro!60}\textbf{8.246} & \cellcolor{Gainsboro!60}\textbf{0.2301} \\
        Hyperprior-64 & & 36.48 & 0.9943 & 15.15 & 0.8135 & \uline{7.118} & \uline{0.3608} & \cellcolor{Gainsboro!60}\textbf{8.031} & \cellcolor{Gainsboro!60}\textbf{0.4527} \\

        Factorized-32 & & 33.29 & 0.9901 & 19.90 & 0.9020 & \uline{7.614} & \uline{0.4443} & \cellcolor{Gainsboro!60}\textbf{12.28} & \cellcolor{Gainsboro!60}\textbf{0.4570} \\
        Factorized-64 & &  33.29 & 0.9901 & 14.99 & 0.8221 & \uline{5.946} & \uline{0.2868} & \cellcolor{Gainsboro!60}\textbf{9.050} & \cellcolor{Gainsboro!60}\textbf{0.5353} \\
        
        Joint-32 & & 35.21 & 0.9911 & 21.38 & 0.9205 & \uline{5.921} & \uline{0.1633} & \cellcolor{Gainsboro!60}\textbf{15.46} & \cellcolor{Gainsboro!60}\textbf{0.7572} \\
        Joint-64 & & 35.21 & 0.9911 & 17.52 & 0.8662 & \uline{5.733} & \uline{0.1335} & \cellcolor{Gainsboro!60}\textbf{11.78} & \cellcolor{Gainsboro!60}\textbf{0.7327} \\

        \bottomrule
      \end{tabular}
  }
    \caption{Average PSNR and MS-SSIM of the reconstructed original images, adversarial examples (AE), and reconstructed adversarial examples across 24 images from the Kodak dataset. The bold number denotes the average degradation of PSNR and MS-SSIM between the adversarial examples and their reconstructed counterparts by our proposed attack method.}
  \label{tab:main_results}
\end{table*}

\textbf{Results for Reconstruction Distortion.}
We generate adversarial examples on the four LIC models with perturbation budget $\epsilon=\{32, 64\}$ and visual results are presented in Fig. \ref{fig:main_result}, which includes 8 images from the Kodak dataset. As observed, the LIC successfully reconstructs the original image. However, for the adversarial image, despite the perturbation being almost imperceptible and unnoticeable, the reconstructed adversarial examples are significantly disrupted. Table \ref{tab:main_results} presents the average PSNR and MS-SSIM of the reconstructed original images, adversarial examples, and reconstructed adversarial examples across 24 Kodak images.
As shown in the Table, the average PSNR and MS-SSIM of reconstructed adversarial examples, denoted by the number with the underline, are extremely low, which verifies the effectiveness of our proposed attack. We also evaluate the average degradation of PSNR and MS-SSIM measured between the adversarial examples and their reconstructed counterparts, marked as bold. Notably, the attack on the \texttt{Joint} model with $\epsilon=32$ can achieve an MS-SSIM degradation of 0.7572 (decreased from 0.9205 to 0.1633). This demonstrates that the reconstructed images are almost completely destroyed.

\begin{figure*}[t]
	\centering 
	\includegraphics[width=\linewidth]{ 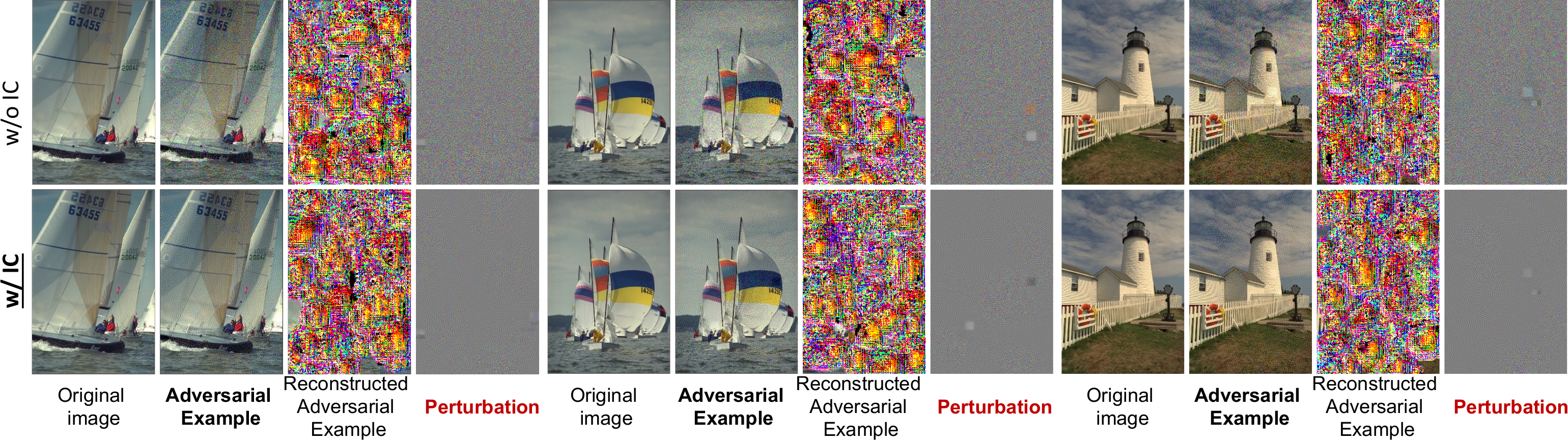}
	\caption{Impact of proposed IC. The visualization depicts perturbations generated from attacks with or without IC. As shown in the figure, the adversarial example generated with IC exhibits a more natural behavior compared to those without IC.}
	\label{fig:effect}
\end{figure*}

\begin{table}[t]
\setlength{\tabcolsep}{8pt}
  \centering
  \scalebox{1}{
      \begin{tabular}{cccccc}
        \toprule
        & $\epsilon$ & Low & Middle & High & Average \\
        \midrule
        w/o IC & 32 & 0.8268 & 0.8359 & 0.8374 & 0.8333 \\
        \rowcolor{Gainsboro!60} \textbf{w/ IC} & 32 & \textbf{0.9315} & \textbf{0.9015} & \textbf{0.9006} & \textbf{0.9112} \\
        \midrule 
        w/o IC & 64 & 0.6820 & 0.6781 & 0.6873 & 0.6824 \\
        \rowcolor{Gainsboro!60} \textbf{w/ IC} & 64 & \textbf{0.8570} & \textbf{0.8147} & \textbf{0.8126} & \textbf{0.8281} \\

        \bottomrule
      \end{tabular}
  }
    \caption{The MS-SSIM of adversarial examples compared to original images on the low, middle, and high quality \texttt{Anchor} models with or without high-frequency constraints. A higher MS-SSIM indicates that the adversarial perturbations are more imperceptible. }
  \label{tab:effectIC}
\end{table}

\textbf{Effect of High-frequency Perturbation.}
We evaluate the impact of our proposed IC on the \texttt{Anchor} model, with a perturbation budget of $\epsilon=\{32, 64\}$. The low, middle, and high models are trained with quality level coefficients $\lambda$ 0.0130, 0.0250, and 0.0483, respectively.
The results of our proposed attack on the high-quality model with $\epsilon=32$ in Fig. \ref{fig:effect}, which contains three Kodak images. Table \ref{tab:main_results} presents the average MS-SSIM of the adversarial examples across 24 images. As observed in the column of adversarial examples and perturbations, LIC successfully disrupts the reconstructed adversarial examples, while the adversarial examples generated by our proposed IC achieve imperceptible perturbations. In contrast, adversarial examples without the IC show conspicuous noises. As shown in Table \ref{tab:effectIC}, an attack with IC can increase the average MS-SSIM by 0.078 and 0.146, respectively, demonstrating a substantial difference in image reconstruction quality.

\begin{figure}[ht]
	\centering 
	\includegraphics[width=0.95\linewidth]{ 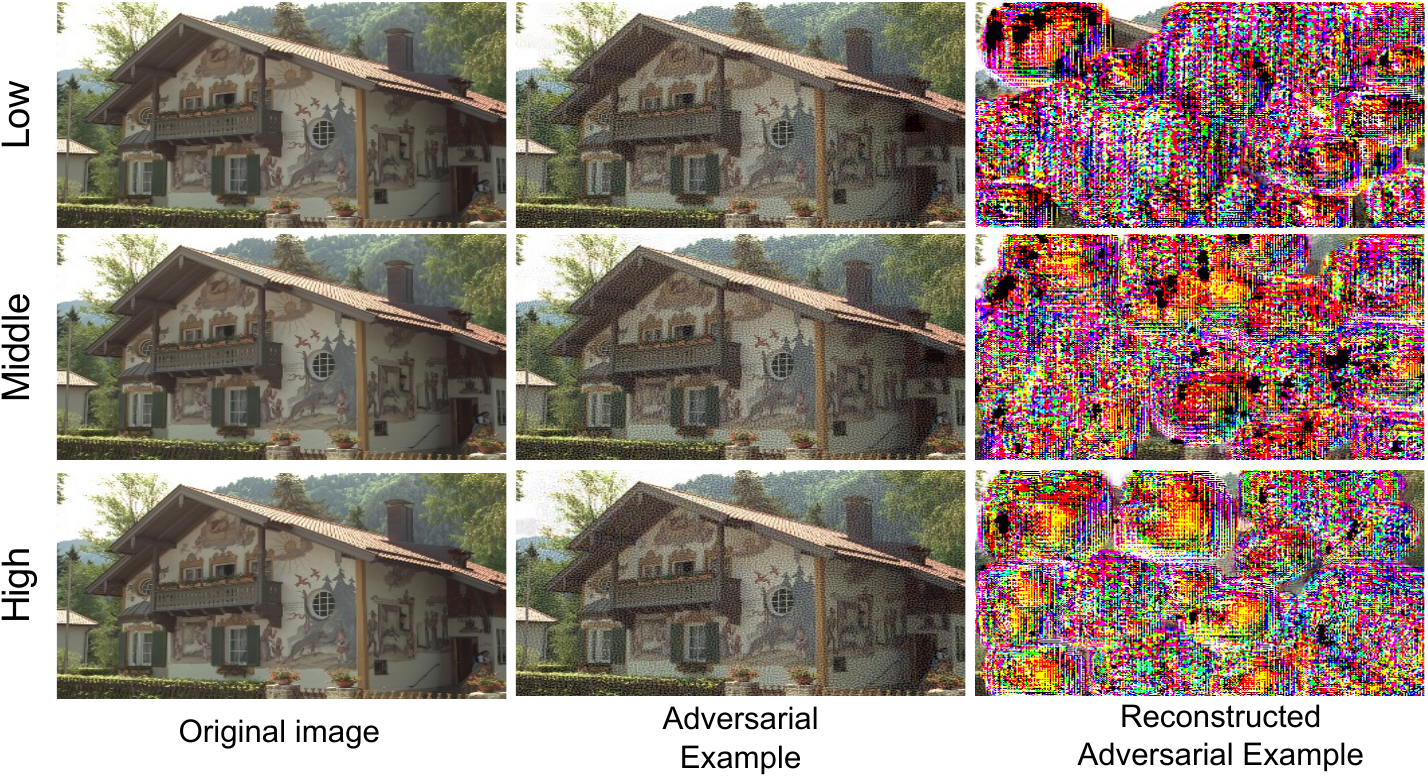}
	\caption{Perturbation of different quality levels.}
	\label{fig:difq}
\end{figure}

\begin{table*}[t]
\setlength{\tabcolsep}{1pt}
  \centering
  \scalebox{0.85}{
      
      \begin{tabular}{cccccccccc}
        \toprule
          \multirow{3}{*}{\makecell{ Reconstruction \\  Quality }} & & \multicolumn{2}{c}{Reconstructed Original Image} & \multicolumn{2}{c}{AE} & \multicolumn{2}{c}{Reconstructed AE} & \multicolumn{2}{c}{$\Delta$ of AE vs. Reconstructed AE} \\
        
        \cmidrule{3-4} \cmidrule{5-6} \cmidrule{7-8} \cmidrule{9-10}
        & & PSNR & MS-SSIM & PSNR & MS-SSIM & PSNR & MS-SSIM & $\downarrow$ PSNR & $\downarrow$ MS-SSIM \\
        \hline
        \multicolumn{10}{c}{Anchor} \\ \hline
        Low & & 32.01 & 0.9838 & 23.96 & 0.9315 & \uline{11.57} & \uline{0.4006} & \cellcolor{Gainsboro!60} \textbf{12.39} & \cellcolor{Gainsboro!60} \textbf{0.5309} \\
        Middle & & 33.16 & 0.989  & 21.68 & 0.9015 & \uline{5.510}  & \uline{0.1274} & \cellcolor{Gainsboro!60} \textbf{16.17} & \cellcolor{Gainsboro!60} \textbf{0.7741} \\
        High & & 33.81 & 0.9914 & 21.78 & 0.9006 & \uline{5.421} & \uline{0.1285} & \cellcolor{Gainsboro!60} \textbf{16.35} & \cellcolor{Gainsboro!60} \textbf{0.7721} \\
        Average & & 32.99 & 0.9880 & 22.47 & 0.9112 & \uline{7.500} & \uline{0.2188} & \cellcolor{Gainsboro!60} \textbf{14.97} & \cellcolor{Gainsboro!60} \textbf{0.6923} \\
        \midrule
        \multicolumn{10}{c}{Factorized} \\ \hline
        Low & & 29.97 & 0.9831 & 20.36 & 0.9002 & \uline{7.786} & \uline{0.4178} & \cellcolor{Gainsboro!60} \textbf{12.57} & \cellcolor{Gainsboro!60} \textbf{0.4824} \\
        Middle & & 34.07 & 0.9929 & 19.76 & 0.8945 & \uline{8.270} & \uline{0.4779} & \cellcolor{Gainsboro!60} \textbf{11.49} & \cellcolor{Gainsboro!60} \textbf{0.4166} \\
        High & & 35.84 & 0.9944 & 19.59 & 0.9115 & \uline{6.787} & \uline{0.4374} & \cellcolor{Gainsboro!60} \textbf{12.80} & \cellcolor{Gainsboro!60} \textbf{0.4741} \\
        Average & & 33.29 & 0.9901 & 19.90 & 0.9020 & \uline{7.614} & \uline{0.4443} & \cellcolor{Gainsboro!60} \textbf{12.28} & \cellcolor{Gainsboro!60} \textbf{0.4577} \\

        \bottomrule
      \end{tabular}
  }
    \caption{Results on LIC models with different quality levels.}
  \label{tab:cheng2020_dif_quality1}
\end{table*}

\textbf{Effect on LIC Models with Different Quality Levels.} We further evaluate our proposed attack method on \texttt{Anchor} and \texttt{Factorized} LIC models across low, middle, and high reconstruction quality levels.
Table \ref{tab:cheng2020_dif_quality1} presents the results with varying quality levels. For example, for the low, middle, and high-quality models of \texttt{Factorized}, our proposed method consistently achieves an MS-SSIM degradation of 0.4824, 0.4166, and 0.4741, respectively, averaging 0.4577. The results illustrate that the proposed reconstruction distortion can affect all quality levels.

\textbf{Findings.} Our experiments have led to several intriguing observations.
\ding{182} Besides arbitrary noises, the generated adversarial perturbation also contain certain irregular patterns. For instance, it can be observed in Fig. \ref{fig:effect} that there are small square patterns within each generated perturbation. We hypothesize that these specific areas may have a significant impact on reconstruction quality. Future work may investigate designing countermeasures for detecting and defending the adversarial attack leveraging these patterns.
\ding{183} Different LIC models demonstrate varying levels of robustness. From Fig. \ref{fig:main_result}, we find that $\texttt{Hyperprior}$ and $\texttt{Joint}$ appear to be more robust than others.
Based on this, we hypothesize that LIC models with higher-quality reconstruction capabilities also have superior robustness.



\section{Conclusion}
In this paper, we explore the robustness of LIC by launching adversarial quality attacks based on the Frobenius norm-based loss function to create adversarial examples that maximize the deviation between the original and the reconstructed images and introduce the IC to ensure the perturbations are invisible to human perception. Experiments on the Kodak dataset and various LIC models illustrate the effectiveness and reveal intriguing findings, including irregular perturbation patterns and varying levels of robustness across different LIC models.

\normalem
{
\bibliographystyle{plain}
\bibliography{ref}
}


\end{document}